%% file: main.tex
\documentclass[10pt,twocolumn,letterpaper]{article}

\usepackage[pagenumbers]{wacv} 


\definecolor{wacvblue}{rgb}{0.21,0.49,0.74}
\usepackage[pagebackref,breaklinks,colorlinks,allcolors=wacvblue]{hyperref}

\def\wacvPaperID{212} 
\def\confName{WACV}
\def\confYear{2026}

\usepackage{graphicx}
\usepackage{amsmath}
\usepackage{amssymb}
\usepackage{booktabs}
\usepackage{tikz}
\usepackage{multirow}
\usepackage{booktabs}
\usepackage{colortbl} 
\usepackage{xcolor} 
\definecolor{lightred}{rgb}{1.0, 0.94, 0.94}
\usepackage{amsmath}
\usepackage{algorithm}
\usepackage{algpseudocode}
\usepackage{makecell}
\usepackage{bm}
\usepackage{tikz}
\usepackage{amsmath}
\usetikzlibrary{positioning, shapes.geometric, fit}
\usepackage{pifont}
\usepackage{hhline}
\usepackage{float}
\usepackage{placeins} 

\title{PSA-MIL: A Probabilistic Spatial Attention-Based Multiple Instance Learning for Whole Slide Image Classification}

\author{Sharon Peled \quad  Yosef E. Maruvka\thanks{These authors are co-senior authors.} \quad  Moti Freiman\footnotemark[1]\\
Technion – Israel Institute of Technology, Haifa, Israel.\\
{\tt\small sharonpe@campus.technion.ac.il}
}

\begin{document}
\maketitle


\begin{abstract}
Whole Slide Images (WSIs) are high-resolution digital scans widely used in medical diagnostics. Due to their immense size, WSI classification is typically approached using Multiple Instance Learning (MIL), where a slide is partitioned into individual tiles, disrupting its spatial structure.
Recent MIL methods often incorporate spatial context through rigid spatial assumptions (e.g. fixed kernels), which limit their ability to capture the intricate tissue structures crucial for an accurate diagnosis.
To address this limitation, we propose \textbf{P}robabilistic \textbf{S}patial \textbf{A}ttention MIL (PSA-MIL), a
novel attention-based MIL framework that integrates spatial context into the attention mechanism through learnable distance-decayed priors, formulated within a probabilistic interpretation of self-attention as a posterior distribution.    
This formulation enables a dynamic inference of spatial relationships during training, eliminating the need for predefined assumptions often imposed by previous approaches. 
Furthermore, we introduce a diversity loss that promotes complementary spatial representations across attention heads and a spatial posterior-pruning strategy that reduces computational cost for long WSI sequences while preserving performance. 
Extensive experiments across multiple datasets and tasks show that PSA-MIL outperforms current baselines and achieves state-of-the-art results with substantially lower computational overhead.
Our code is available at \url{https://github.com/SharonPeled/PSA-MIL}.


\end{abstract}

\begin{figure}[ht]
\centering
\includegraphics[width=0.7\columnwidth]{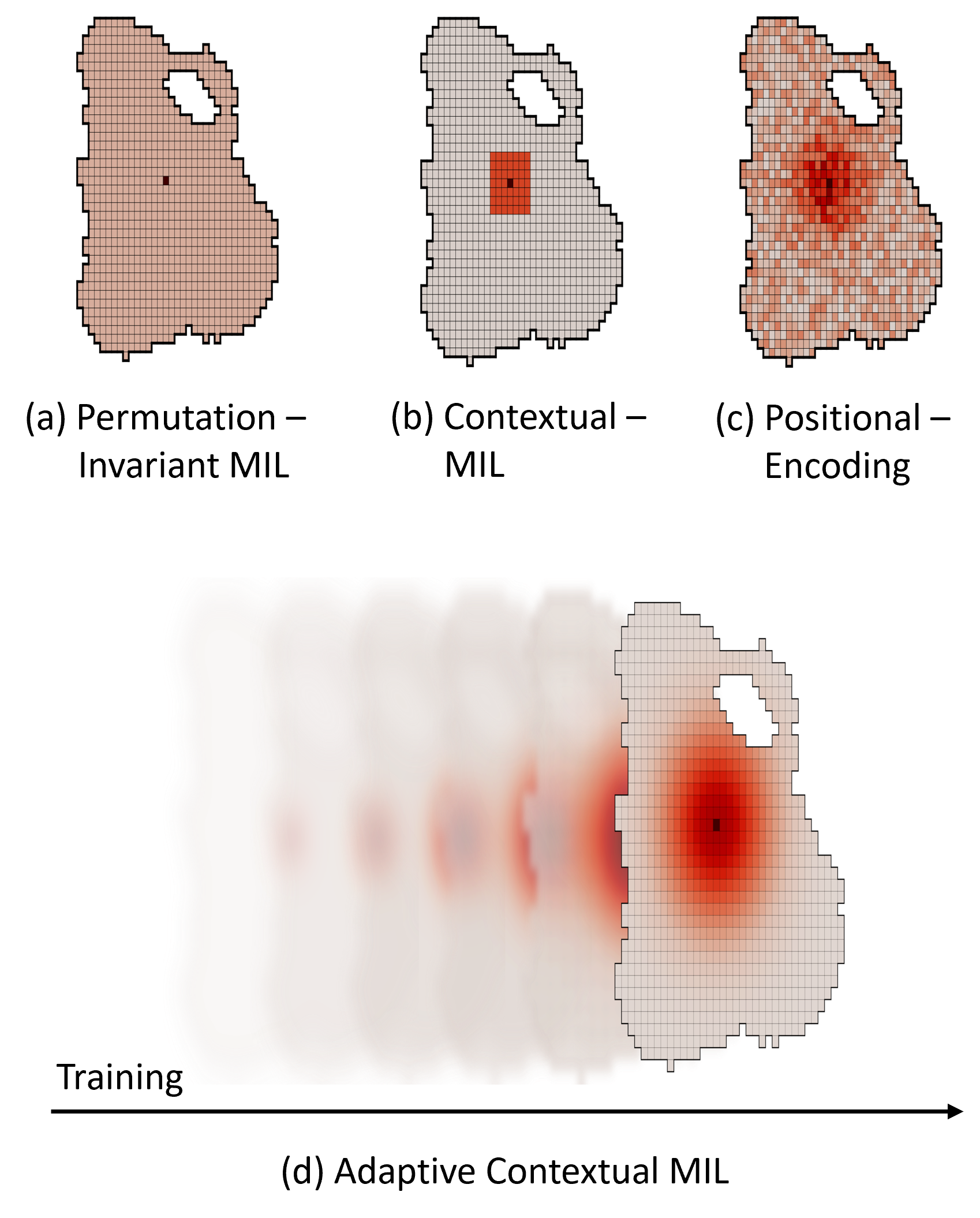}
\caption{Comparison of spatial context modeling approaches in WSI analysis.
(a) Permutation-invariant MIL treats all patches equally, disregarding spatial relationships.
(b) Contextual MIL restricts interactions to a predefined neighborhood around the anchor patch, enforcing a fixed spatial structure.
(c) Positional encoding incorporates spatial bias into tile representations, providing limited spatial awareness. Its unstructured nature lacks alignment with physical distances, which hurts interpretability.
(d) Our proposed PSA-MIL dynamically learns spatial relationships, starting with an initial distribution over neighboring patches and evolving throughout training, in contrast to previous methods that maintain a fixed spatial structure.}
\label{concept_fig}
\end{figure}

\section{Introduction}
Histopathological image analysis plays a crucial role in modern medicine, often regarded as ground truth in medical practice \cite{kumar2020whole, hanna2020whole}. 
The digitization of pathological slides into Whole Slide Images (WSIs) has revolutionized the field of pathology, enabling large-scale computational analysis.
Due to the immense size of a WSI, which can exceed a billion pixels, a widely adopted approach for computational analysis is to decompose these large images into smaller, processable patches. Exhibiting only slide-level annotations, histopathological image analysis is commonly formulated as a Multiple Instance Learning (MIL) task \cite{gadermayr2024multiple}, where each slide is modeled as a permutation-invariant bag of instances, overlooking spatial context among instances.

\noindent Recognizing the critical role of spatial understanding of tissue structures during histopathological examination \cite{ruusuvuori2022spatial, levy2020spatial, huang2024qust}, recent advancements have introduced \textit{contextual MIL} approaches which incorporate spatial relationships into the learning process \cite{fourkioti2023camil, bontempo2023mil,hou2022h, cui2023bayes}.
While contextual models have shown strong performance, they often struggle to fully capture spatial dependencies due to fixed spatial assumptions. 
For example, CAMIL \cite{fourkioti2023camil} imposes predefined spatial constraints via a neighbor-constrained attention module, whereas Bayes-MIL \cite{cui2023bayes} enforces spatial smoothness using a CRF with a fixed kernel size, both limiting flexibility in capturing diverse spatial patterns.
Others, such as DAS-MIL\cite{bontempo2023mil} and H²-MIL\cite{hou2022h}, focus on distilling multi-scale and hierarchical representations.
\noindent Another popular approach for incorporating spatial context is positional encoding and its many variants \cite{zhao2022setmil, li2024rethinking, chen2022scaling, shao2021transmil}.
For example, SETMIL \cite{zhao2022setmil} employs Relative Positional Encoding (RPE) to model relative positions, and TransMIL \cite{shao2021transmil} leverages PPEG, a convolution-based implicit positional encoding.
However, our experiments indicate that their improvements are often marginal, and their integration of spatial context is primarily empirically justified rather than theoretically grounded, resulting in a lack of a formal interpretative framework.

\noindent Furthermore, long sequences, such as those in WSI analysis, pose a challenge for transformer-based architectures due to the quadratic complexity of self-attention. Solutions like low-rank and kernel-based approximations, including Nyströmformer \cite{xiong2021nystromformer} used in TransMIL \cite{shao2021transmil} and in CAMIL \cite{fourkioti2023camil}, often deliver suboptimal performance \cite{dao2022flashattention,han2023flatten}. 
Other methods use local attention to reduce computations \cite{li2024rethinking}, but are limited to fixed spatial context.

\noindent To this end, we propose \textbf{P}robabilistic \textbf{S}patial \textbf{A}ttention MIL (PSA-MIL), a novel attention-based MIL approach that seamlessly integrates spatial context into the attention mechanism. 
Our method builds upon a probabilistic interpretation of self-attention, incorporating spatial relationships as learnable distance-decayed priors within a posterior distribution. 
This incorporation enables the model to capture how instances attend to each other based on their spatial proximity, balancing adaptive priors with observed likelihood.
\cref{concept_fig} compares existing spatial modeling approaches with our proposed PSA-MIL.
The main contributions of this paper can be summarized as follows:
\begin{itemize}

    \item \textbf{A Principled Self-Attention Formulation for Adaptive Contextual MIL:} We introduce a probabilistic self-attention framework that learns spatial dependencies dynamically during training. 
    This data-driven approach enhances spatial context modeling in WSI analysis, overcoming the limitations of prior methods that rely on fixed assumptions and are often subject to extensive tuning. 
    
    \item \textbf{Dynamic Multi-Head Spatial Specialization:}
    Our formulation enables each attention head to independently derive its own spatial interaction range — a property unique to our approach.
    
    \item \textbf{An Entropy-based Diversity Loss for Multi-head Attention:} We introduce a loss term that explicitly encourages spatial variations among attention heads, ensuring that each head captures distinct spatial representations. This mitigates redundant representations often observed in multi-head architectures \cite{michel2019sixteen,zhang2021enlivening,bian2021attention}.
    
    \item \textbf{A Spatial Pruning Strategy for Computational Efficiency:} 
    We propose a spatial pruning strategy for the posterior estimation that significantly reduces the quadratic complexity of self-attention while preserving critical spatial dependencies. This approach achieves substantial computational savings without compromising performance, in contrast to other low-cost self-attention approximations that may yield suboptimal solutions \cite{li2024rethinking, dao2022flashattention, han2023flatten}.
    

\end{itemize}




\section{Related Work}
\subsection{Multiple Instance Learning in WSI Analysis}
Multiple Instance Learning (MIL) has become a widely adopted approach for WSI classification, allowing models to make slide-level predictions from weakly supervised annotations by treating whole slides as bags of image tiles. \cite{castro2024sm, cui2023bayes, zhang2022dtfd, bontempo2023graph}. 
\noindent Attention-based Multi-Instance Learning (ABMIL) allowed to learn instance-level importance weights \cite{ilse2018attention}.
Several works have since refined feature aggregation. For example, CLAM \cite{lu2021data} incorporated a prototype-based clustering mechanism, and DTFD-MIL \cite{zhang2022dtfd} employed a multi-tier fusion strategy.
\noindent More recently, Transformer-based MIL architectures have emerged, leveraging self-attention to capture long-range dependencies between tiles \cite{shao2021transmil, fourkioti2023camil, castro2024sm, zhao2022setmil}. 
For example, TransMIL \cite{shao2021transmil} modeled global and local features via token interactions, while GTP \cite{zheng2022graph} combined GNNs and Transformers for topology-aware spatial modeling.

\subsection{Spatial Context in Multiple Instance Learning}
Reflecting the importance of spatial understanding in histopathology \cite{ruusuvuori2022spatial,levy2020spatial,huang2024qust}, several approaches have been proposed for modeling spatial context in MIL. 
Graph-based approaches model spatial relationships by treating tiles as graph nodes with structured dependencies \cite{castro2024sm, zheng2022graph, bontempo2023graph}.
For example, SM-MIL \cite{castro2024sm} applies a smoothness operator, enforcing gradual transitions in attention scores to better capture local dependencies.
Although effective, these methods are prune to over-smoothing, limiting the model’s ability to capture fine-grained spatial distinctions \cite{oono2019graph, chen2020measuring}. 
Additional analysis and comparisons of GNN-based spatial modeling are provided in the supplementary material.
Bayes-MIL \cite{cui2023bayes} takes a different approach by incorporating spatial context through a Convolutional CRF, however, it still relies on a fixed kernel size, which restricts its ability to adapt.

\noindent Another direction for modeling spatial context involves positional encoding, which embeds spatial bias into tile representations to enhance spatial awareness \cite{li2024rethinking, zhao2022setmil, chen2022scaling, shao2021transmil}. For example, TransMIL \cite{shao2021transmil} employs PPEG, a convolution-based positional encoding, while SETMIL \cite{zhao2022setmil} utilizes Relative Positional Encoding (RPE) to model spatial relationships.
While proven effective, our experiments show that their performance gains are often marginal. 
More recent approaches utilize spatially constrained Transformer-based MIL architectures, where interactions are restricted to fixed neighborhoods. For example, CAMIL \cite{fourkioti2023camil} enforces predefined spatial constraints through a neighbor-constrained attention mechanism, while LongMIL \cite{li2024rethinking} applies a fixed attention mask to model local interactions, both limiting flexibility in capturing diverse spatial patterns.
Moreover, due to the quadratic complexity of self-attention, Transformer-based MIL architectures \cite{shao2021transmil, fourkioti2023camil} often rely on self-attention approximations, such as Nyströmformer \cite{xiong2021nystromformer}, which can lead to suboptimal performance \cite{dao2022flashattention, han2023flatten}.
In contrast, our PSA-MIL approach dynamically infers spatial relationships during training, adapting to the underlying tissue structure without rigid constraints. This paradigm shift is illustrated in \cref{concept_fig}.
Additionally, our spatial pruning strategy eliminates the need for self-attention approximations, ensuring efficiency without compromising performance.



\begin{figure*}[ht]
\centering
\includegraphics[width=0.9\textwidth]{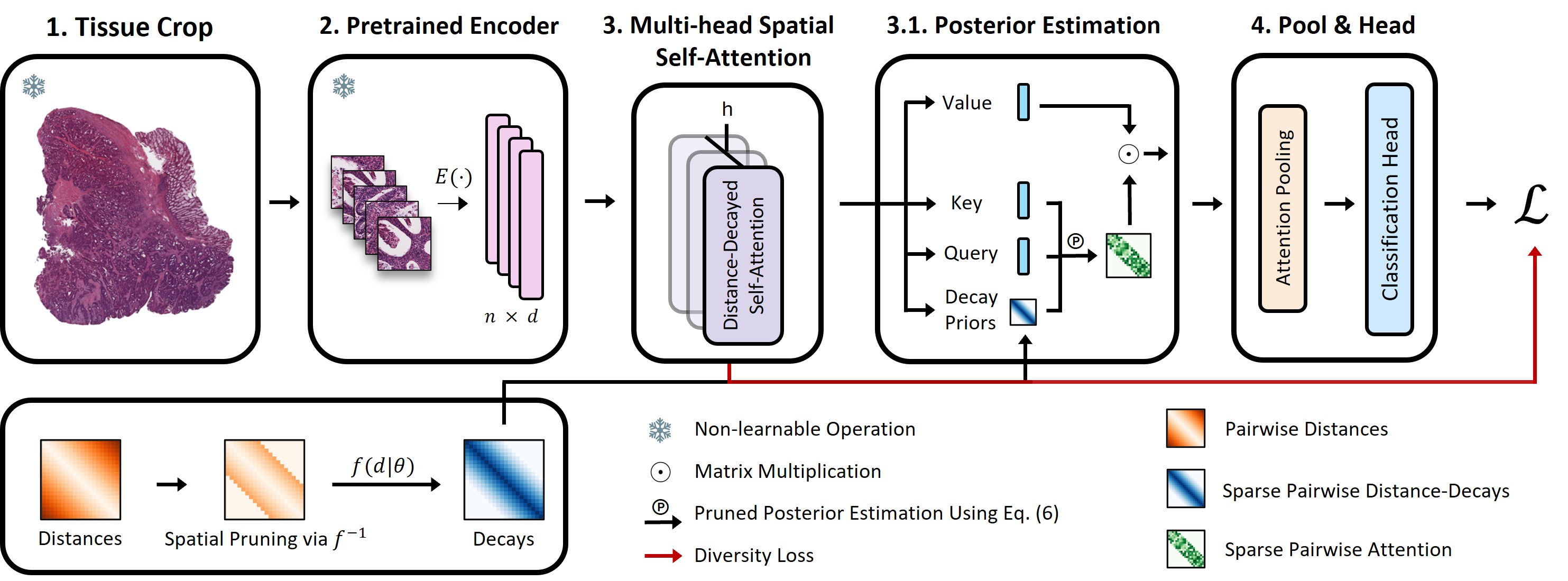}
\caption{PSA-MIL Overview: 
1–2. Tissue regions are cropped and encoded using a pretrained feature extractor, producing tile representations.
3-3.1. A multi-head spatial self-attention mechanism is employed to generate informative, spatially correlated tile representations.
Within this block, the pairwise distance matrix (bottom left) undergoes dynamic spatial pruning (\cref{spatial_prune}) via the learned function \( f^{-1}(\tau|\theta) \), producing a spatially pruned distance matrix. The distance-decayed matrix is then computed using \( f(d|\theta) \). Both \( f^{-1}(\tau|\theta) \) and \( f(d|\theta) \) are learnable components parameterized by \( \theta \), enabling adaptive pruning and decay.
Simultaneously, query, key, and value matrices are generated. The posterior attention distribution is then computed using \cref{final_posterior} and applied to the value matrix, producing refined tile embeddings that encode spatial interactions.
4. These embeddings are subsequently pooled via an attention-based aggregation mechanism and passed through a classification head to generate slide-level predictions. 
The final loss function consists of both the classification loss and a diversity loss (\cref{diversity}) applied to the multi-head attention mechanism to encourage the capture of diverse spatial patterns.
}
\label{main_fig}
\end{figure*}

\section{Methodology}

\subsection{Background} 
\textbf{Preliminaries} \quad Given a WSI, tissue regions are cropped into tiles and encoded using a pretrained frozen encoder \( E(\cdot) \), forming an instance embeddings matrix \( X \in \mathbb{R}^{n \times d} \). 



\subsubsection{Self-Attention for Multiple Instance Learning}
\label{self_attention_MIL}



Under the MIL framework, self-attention enables each instance (or tile) to interact with all other instances in the bag. The attention is computed as:
\begin{equation} H = \text{softmax}\left(QK^\top / \sqrt{d_k}\right)V 
\label{eq:sa}
\end{equation}

\noindent where \( Q \), \( K \), and \( V \) are linear projections of the instance embeddings and \(d_k\) denotes the dimension of Keys.

\noindent \textbf{Multi-head Attention:} Multiple attention heads are concatenated and projected to produce the final output:

\[
\text{Multi-Head}(\{Q, K, V\}_{i=1}^h) = \text{Concat}(H_1, \ldots, H_h)W
\]

\noindent The resulting attention outputs are aggregated via a pooling operator to compute the slide prediction. 

\noindent While this approach captures intra-bag dependencies, it ignores spatial relationships among instances, which may limit the model's ability to capture intricate structures.








\subsubsection{Self-Attention for Multiple Instance Learning as Posterior Distribution}  
\label{attention_posterior}
In the context of MIL, self-attention mechanism can be interpreted as computing posterior distributions of each tile's query vector $q_i$ over all key vectors $k_j$, effectively modeling the attention process as a Gaussian Mixture Model (GMM) \cite{nguyen2022improving, nguyen2022improving2}.
Consider the query vector \( q_i \in \mathbb{R}^{d_k} \) for tile \( i \) and the key vectors \( k_j \in \mathbb{R}^{d_k} \) for tiles \( j = 1, \dots, N \). Let \( t \in \{0,1\}^N \) be a one-hot encoded vector indicating the selection of key \( k_j \). The distribution of the query vector \( q_i \) is modeled as:

\vspace{-6pt}  

\begin{equation}
p(q_i) = \sum_{j=1}^N\! \pi_j p(q_i \mid t_j = 1) = \sum_{j=1}^N\! \pi_j \mathcal{N}(q_i \mid k_j, \sigma^2 I)
\end{equation}

\noindent where $\pi_j = p(t_j = 1)$ is the prior probability, and $\mathcal{N}(q_i \mid k_j, \sigma^2 I)$ denotes a Gaussian distribution with mean $k_j$ and covariance $\sigma^2 I$.

\noindent The posterior probability that $q_i$ corresponds to $k_j$ is given by:


\vspace{-2pt}  

\begin{align}
&p(t_j = 1 \mid q_i) = \dfrac{\pi_j \mathcal{N}(q_i \mid k_j, \sigma^2 I)}{\sum_{j'=1}^N \pi_{j'} \mathcal{N}(q_i \mid k_{j'}, \sigma^2 I)} \label{eq:posterior}  \\[5pt]
&= \dfrac{\pi_j \exp \left[-(\|q_i\|^2 + \|k_j\|^2) / 2\sigma^2\right] \exp \left(q_i^T k_j / \sigma^2\right)}
{\sum_{j'} \pi_{j'} \exp \left[-(\|q_i\|^2 + \|k_{j'}\|^2) / 2\sigma^2 \right] \exp \left(q_i^T k_{j'} / \sigma^2\right)} \notag 
\end{align}

\vspace{0.5em}

\noindent The self-attention operation described in \cref{eq:sa} implies the following assumptions: 

\vspace{0.5em}

\noindent\textbf{Assumption 1.} \textit{The queries and keys are $\ell_2$-normalized, i.e., $|| q_i || = || k_j || = 1$ for all $i$ and $j$.
} 

\vspace{0.3em}

\noindent\textbf{Assumption 2.} \textit{The prior probabilities are uniform, i.e., $\pi_j = \frac{1}{N}$ for all $j$.}

\vspace{0.3em}

\noindent\textbf{Assumption 3.} \textit{The variance $\sigma^2$ is constant and set to $\sqrt{d_k}$ for all components. }

\vspace{0.5em}

\noindent Under these assumptions, the posterior probability simplifies to:

\vspace{-6pt}  

\begin{equation}
p(t_j = 1 \mid q_i) = \dfrac{\exp\left( q_i^\top k_j / \sqrt{d_k} \right)}{\sum_{j'=1}^N \exp\left( q_i^\top k_{j'} / \sqrt{d_k} \right)}
\end{equation}

\noindent Thus, the posterior distribution $p(t_j = 1 \mid q_i)$ corresponds exactly to the attention weights computed by the standard self-attention, as in \cref{eq:sa}.

\subsection{Learnable Probabilistic Spatial Attention for Adaptive Contextual MIL}  
The self-attention MIL, as formulated in \cref{attention_posterior}, inherently disregards spatial relationships among instances, which is crucial for WSI analysis, where instances represent contiguous regions of tissue and their spatial arrangement provides valuable diagnostic cues \cite{ruusuvuori2022spatial, levy2020spatial, huang2024qust}.
To this end, we propose PSA-MIL, an innovative approach that seamlessly integrates spatial context into the attention mechanism as a learnable component, enabling the model to capture spatial dependencies in a flexible and data-driven manner.
Building upon the probabilistic interpretation of self-attention described in \cref{attention_posterior}, we relax both \textit{Assumption~1 ($\ell_2$-normalized queries and keys)} and \textit{Assumption~2 (uniform priors)}, allowing for a more refined posterior estimation.


\noindent \textbf{Relaxing Assumption~1}  
$\ell_2$-normalization of queries and keys is often introduced to align attention with a probabilistic interpretation \cite{nguyen2022improving}. However, standard self-attention does not inherently enforce this assumption, leading to inconsistencies in the formulation.
We relax this assumption altogether and instead compute the full $\ell_2$-norm 
$|| q_i - k_j ||^2$, allowing queries and keys to retain magnitude information, which can enhance expressiveness \cite{nguyen2022improving}.

\noindent \textbf{Relaxing Assumption~2}  
We relaxed Assumption~2 by replacing the uniform priors with learnable, distance-decayed priors, enabling the capture of spatial relationships among instances. These priors are modeled as parametric, learnable functions, providing a structured and interpretable framework for spatial context.


\noindent \textbf{Pooling Operator} We adopt attention-based pooling due to its effectiveness in aggregating instance-level features and its interpretability. This is a common strategy in WSI analysis \cite{fourkioti2023camil, castro2024sm}. Additional details regarding our architecture design are provided in the supplementary material.

\noindent \cref{main_fig} depicts the complete PSA-MIL pipeline for spatial context modeling in WSI analysis.

\subsubsection{Refined Posterior After Assumption Relaxation}
\label{incorporation}

\noindent Consider a set of instance embeddings \( X \in \mathbb{R}^{n \times d} \), along with their corresponding pairwise distances \( D \in \mathbb{R}^{n \times n} \), where \( d_{ij} \) represents the Euclidean distance between tiles \( i \) and \( j \). 
We define a learnable distance decay function \( f(\cdot|\theta) \), where \( \theta \) is learnable, and \( f(d_{ij}|\theta) \) represents the decay for tile \( j \) from tile \( i \). 

\noindent Utilizing \cref{eq:posterior} for the posterior of \( q_i \), we replace the uniform priors with spatially-informed priors assigned according to \( f(\cdot|\theta) \):
\begin{equation}
    \pi_j = p(t_j = 1) = f(d_{ij}|\theta)
    \label{pie}
\end{equation}
and are learned during training.

\noindent For numerical stability, we incorporate these priors into the posterior distribution by embedding them directly within the exponentials. The overall posterior is then given by:


\begin{equation} 
{\textstyle
p(t_j\! =\! 1 \!\mid\! q_i) \!=\! 
\frac{\exp\left( \dfrac{\|q_i - k_j\|^2}{\sqrt{d_k}} + \log(f(d_{ij}|\theta)) \right)}
{\sum_{j'=1}^N\! \exp\left( \dfrac{\|q_i - k_{j'}\|^2}{\sqrt{d_k}} + \log(f(d_{ij'}|\theta)) \right)}
}
\label{final_posterior}
\end{equation}

\noindent This formulation enables the model to derive the spatial importance directly from the spatial correlations in the data, resulting in a more adaptive and context-aware representation. See full pipeline in \cref{main_fig}.

\vspace{0.5em}

\noindent\textbf{Remark.} \textit{\(f(\cdot | \theta)\) can be interpreted as a form of regularization, constraining the model’s expressiveness by limiting information flow from distant instances.}  



\subsubsection{Parametric Distance-Decay Functions for Spatial Modeling}
\label{parametric}

As detailed in \cref{pie}, the prior distribution is defined by the function \( f(\cdot|\theta) \), which directly models the decay of influence over distance. Consequently, \( f(\cdot|\theta) \) should satisfy the following properties:
\vspace{0.3em}
\begin{enumerate}
    \item \text{Non-negativity}: \( f(d_{ij}|\theta) \geq 0 \) for all \( j \), \( f(d_{ii}|\theta) > 0 \).
    \vspace{0.5em} 
    \item \text{Monotonicity}: 
    \(
    f(d_{ij}|\theta) \geq f(d_{ij'}|\theta) \quad \text{if} \quad d_{ij} \leq d_{ij'}.
    \)
\end{enumerate}
\noindent Note that the normalization factor \( \sum_{j=1}^{N} f(d_{ij}|\theta) \) cancels out in \cref{eq:posterior}.

\noindent However, learning \( f(\cdot|\theta) \) directly  is challenging due to the difficulty of enforcing these constraints during training. To address this, we adopt learnable parametric functions that provide a structured way for dynamically modeling spatial distance-decayed prior probabilities. We experiment with three parametric decay functions, each exhibiting distinct attenuation behavior:

\begin{enumerate}
    \item Exponential Decay:
    \(
    f(d|\lambda) = \exp\left( -\lambda d \right)
    \)
    
    \noindent Provides sharper attenuation, useful when influence drops off quickly with distance.
    
    \item Gaussian Decay:
    \(
    f(d|\sigma) = \exp\left( -\frac{d^2}{2\sigma^2} \right)
    \) 
    
    \noindent Suitable for smoothly decreasing influence over distance, effective for modeling gradual spatial dependencies.
    
    \item Cauchy Decay:
    \(
    f(d|\gamma) = \frac{1}{1 + \left( \frac{d}{\gamma} \right)^2}
    \)
    
    \noindent Captures long-range dependencies, allowing distant tiles to retain influence due to its heavy-tailed nature.
\end{enumerate}




\noindent These parametric decay functions provide a structured, learnable mechanism for modeling spatial influence.
This represents a paradigm shift, enabling principled spatial modeling without imposing predefined constraints, as seen in previous approaches \cite{fourkioti2023camil, cui2023bayes}.
See \cref{concept_fig} for illustration.

\begin{figure}[ht]
\centering
\includegraphics[width=0.8\columnwidth]{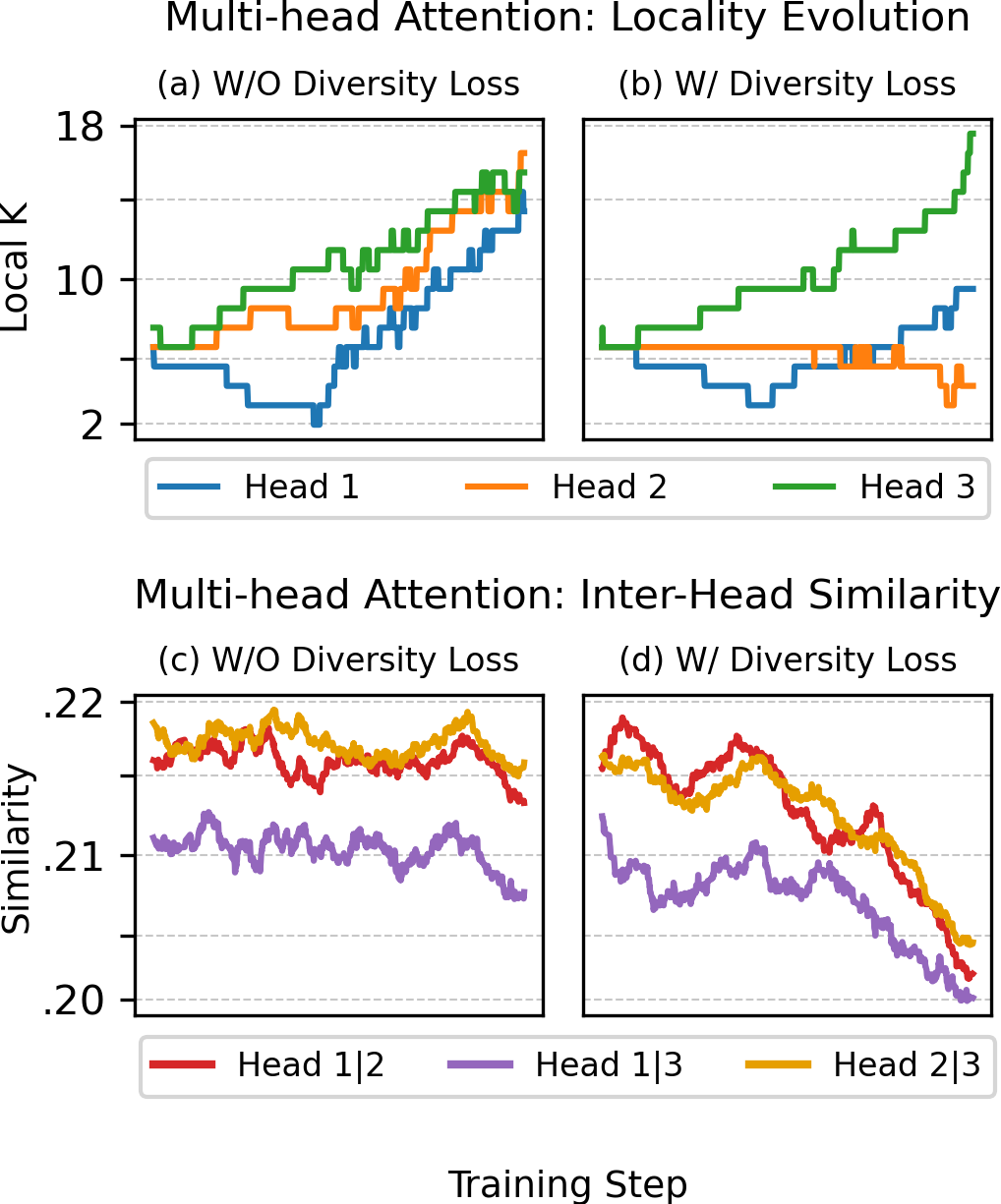}
\caption{\textbf{The Effect of Diversity Loss on Multi-Head Attention.} 
    (a-b). Locality evolution: In \cref{spatial_prune}, we describe how our spatial attention can be interpreted as \textit{dynamic local attention}.
    Without diversity loss, all heads converge to similar locality values, whereas with diversity loss, the locality values diverge.
    (c-d). Inter-head similarity: The similarity metric measures average $\ell_2$ token similarity (smoothed) across heads. Without diversity loss, similarity remains relatively high, while with diversity loss, similarity decreases. 
    See \cref{training_dynamics} for additional details.
    }
\label{div_fig}
\end{figure}

\begin{table*}[t]
    \centering
    \resizebox{\textwidth}{!}{ 

    \renewcommand{\arraystretch}{1.1}
    \begin{tabular}{clcccccc}
        \toprule 
        \multicolumn{2}{c}{Method} & \multicolumn{3}{c}{TCGA-CRC} & \multicolumn{3}{c}{TCGA-STAD} \\
        \cmidrule(lr){3-5} \cmidrule(lr){6-8}
        & & AUC & Accuracy & F1 Score & AUC & Accuracy & F1 Score \\
        \midrule
         \multirow{4}{*}{\rotatebox{90}{\shortstack{Non-\\Contextual}}}
        & ABMIL \cite{ilse2018attention}    & 85.21 $\pm$ 1.01 & 87.18 $\pm$ 1.03 & 62.09 $\pm$ 0.99 & 81.25 $\pm$ 1.45 & 72.98 $\pm$ 0.83 & 84.98 $\pm$ 1.02 \\
        & CLAM \cite{lu2021data}      & 86.18 $\pm$ 0.96 & 87.72 $\pm$ 1.61 & 64.81 $\pm$ 1.35 & 81.71 $\pm$ 0.72 & 76.32 $\pm$ 0.89 & 85.35 $\pm$ 0.72 \\
        & DTFD-MIL \cite{zhang2022dtfd}     & 86.70 $\pm$ 0.94 & 88.01 $\pm$ 1.92 & 66.28 $\pm$ 1.55 & 81.96 $\pm$ 0.32 & 78.12 $\pm$ 0.92 & 85.55 $\pm$ 0.56 \\
        & IBMIL\cite{lin2023interventional}    & 87.51 $\pm$ 1.03 & 88.23 $\pm$ 0.89 & 66.01 $\pm$ 0.56 & 82.31 $\pm$ 1.23 & 78.81 $\pm$ 1.24 & 87.67 $\pm$ 0.89 \\
        \midrule
        \multirow{7}{*}{\rotatebox{90}{Contextual}}
        & TransMIL \cite{shao2021transmil} & 85.94 $\pm$ 1.54 & 87.43 $\pm$ 2.21 & 63.18 $\pm$ 1.23 & 79.97 $\pm$ 1.84 & 75.82 $\pm$ 1.52 & 81.55 $\pm$ 1.11 \\
        & GTP \cite{zheng2022graph}      & 85.87 $\pm$ 2.72 & 88.12 $\pm$ 1.41 & 65.85 $\pm$ 2
        .81 & 80.41 $\pm$ 1.57 & 78.30 $\pm$ 3.61 & 85.47 $\pm$ 2.71 \\
        & SM-MIL \cite{castro2024sm}       & 87.84 $\pm$ 0.45 & 88.86 $\pm$ 0.29 & 64.60 $\pm$ 2.53 & 81.23 $\pm$ 1.37 & \textbf{79.79 $\pm$ 1.88} & 87.85 $\pm$ 0.57 \\
        & BAYES-MIL \cite{cui2023bayes}    & 86.66 $\pm$ 1.23 & 87.57 $\pm$ 0.93 & 66.93 $\pm$ 2.32 & 82.11 $\pm$ 2.12 & 78.41 $\pm$ 0.65 & 87.39 $\pm$ 0.51 \\
        & \cellcolor{lightred} PSA-MIL\textsubscript{[Exp]} &  \cellcolor{lightred} 88.27 $\pm$ 0.94 & \cellcolor{lightred} 88.13 $\pm$ 0.92 & \cellcolor{lightred} 66.29 $\pm$ 1.55 & \cellcolor{lightred} 82.96 $\pm$ 0.92 & \cellcolor{lightred} 78.48 $\pm$ 0.87 & \cellcolor{lightred} 89.28 $\pm$ 1.01 \\
        & \cellcolor{lightred} PSA-MIL\textsubscript{[Gau]} &  \cellcolor{lightred} \textbf{88.97 $\pm$ 1.01} & \cellcolor{lightred} 87.48 $\pm$ 1.03 & \cellcolor{lightred} 66.27 $\pm$ 1.99 & \cellcolor{lightred} 82.98 $\pm$ 1.12 & \cellcolor{lightred} 79.09 $\pm$ 1.83 & \cellcolor{lightred} \textbf{89.53 $\pm$ 1.00} \\
        & \cellcolor{lightred} PSA-MIL\textsubscript{[Cau]} &  \cellcolor{lightred} 87.50 $\pm$ 1.04 & \cellcolor{lightred} \textbf{88.93 $\pm$ 1.21} & \cellcolor{lightred} \textbf{67.70 $\pm$ 1.23} & \cellcolor{lightred} \textbf{83.06 $\pm$ 0.57} & \cellcolor{lightred} 78.59 $\pm$ 0.45 & \cellcolor{lightred} 89.52 $\pm$ 1.31 \\
        \bottomrule
    \end{tabular}}
    \caption{Subtyping performance of different MIL approaches on TCGA-CRC and TCGA-STAD datasets, categorized into Non-Contextual and Contextual methods.
    PSA-MIL variants achieve top performance across both tasks, demonstrating the benefits of our spatial modeling.}
    \label{main_table}
\end{table*}

\subsection{Diversity Loss for Spatial Decay Parameters}
\label{diversity}
Multi-head attention allows the model to learn nuanced representations by having each head focus on different aspects of the input \cite{vaswani2017attention}.
However, multi-head mechanisms are often prone to redundancy and entanglement, which can limit the diversity of learned representations and lead to suboptimal outcomes \cite{michel2019sixteen,zhang2021enlivening,bian2021attention}.
To overcome this issue, we suggest an entropy-based loss term that encourages diversity among \( f(\cdot|\theta) \), enabling the model to capture diverse spatial patterns within the data.

\noindent Let $\theta_h$ be the spatial decay parameter for head $h$, drawn from an unknown distribution $p(\theta)$.  
To approximate $p(\theta)$, we use Kernel Density Estimation (KDE) \cite{chen2017tutorial}:

\vspace{-5pt}

\begin{equation}
    \hat{p}(\theta) = \frac{1}{H\sigma} \sum_{h=1}^{H} K\left( \frac{\theta - \theta_h}{\sigma} \right),
\end{equation}

\noindent where $K(x)$ is a Gaussian kernel with bandwidth $\sigma$.  

\noindent To encourage variation among heads, we introduce an entropy-based loss term that promotes diversity in $\{\theta_h\}_{h=1}^{H}$.
The entropy of $p(\theta)$ is given by: \(H(p) = -\mathbb{E}_{\theta \sim p(\theta)} [\log p(\theta)]\), which we estimate via Monte Carlo: 

\vspace{-6pt}
\begin{equation}
    H(p) \approx -\frac{1}{M} \sum_{m=1}^{M} \log \hat{p}(\tilde{\theta}_m),
\end{equation}

\noindent where $\tilde{\theta}_m$ are samples drawn from $\hat{p}(\theta)$.  

\noindent The entropy-based diversity loss is then defined as \(\mathcal{L}_{\text{Diversity}} = -H(p)\) and incorporated into the final objective:

\begin{equation}
    \mathcal{L} = \mathcal{L}_{\text{CE}} + \alpha \mathcal{L}_{\text{Diversity}},
\end{equation}
\noindent where $\mathcal{L}_{\text{CE}}$ is the cross-entropy loss and \(\alpha\) is a hyperparameter.
\noindent By encouraging high entropy, PSA-MIL achieves a more disentangled and diverse spatial representations. 
\cref{div_fig} demonstrates the effect of the diversity loss on locality convergence and the disentanglement of representations.

\subsection{Spatially Pruning the Posterior}
\label{spatial_prune}
The quadratic complexity of self-attention poses a challenge for long sequences, such as in WSI analysis. Solutions like low-rank and kernel-based approximations often deliver suboptimal performance \cite{dao2022flashattention,han2023flatten}.

\noindent To address this challenge, we propose a spatial pruning strategy for the posterior distribution that restricts attention computations based on learned spatial dependencies.
By leveraging our prior distribution modeling, as described in \cref{parametric}, we are able to identify insignificant interactions \textit{before full computation}, thereby achieving substantial cost reduction without compromising performance.
Namely, we apply a filtering criterion \( f(d|\theta) \geq \tau \), where \( \tau \) is a threshold parameter, permitting only instances with decays above \( \tau \) to contribute to the posterior computation.
Such pruning techniques have been explored in mixture models, demonstrating the effectiveness of removing components that contribute negligibly to the posterior \cite{svensen2005robust, nguyen2023probabilistic}.

\noindent By leveraging an invertible parametric function \( f(d|\theta) \), we derive a guaranteed upper bound on spatial interactions: \( d \leq f^{-1}(\tau | \theta)\). As a result, the computational complexity is reduced to \( O(n \cdot K^2) \ll O(n^2) \), where \(K=\lceil f^{-1}(\tau | \theta) \rceil\).
This yields a \textit{dynamic local attention}, where \(K\) is learned via \(\theta\), and the retained instances are weighted by their learned decay values, unlike fixed local attention.
\cref{div_fig} (a-b) shows the evolution of our dynamic local attention during training.
To the best of our knowledge, this is the first instance where an attention locality is derived during training in an end-to-end fashion.
Notably, we do not fine-tune \(\tau\); we typically set it to a constant value (e.g., \(1e{-3}\)), which we find works well in practice.
\cref{main_fig} illustrates how \(f^{-1}(\tau | \theta)\) is applied in the PSA-MIL pipeline.

\section{Experiments}\label{experiments}
We conducted extensive experiments spanning cancer subtyping, metastatic detection, survival prediction, and patch-level localization, with further analyses of computational efficiency, training dynamics, and attention visualizations.
Unless otherwise specified, the Lunit foundation model \cite{kang2023benchmarking} was used as our pretrained encoder. An NVIDIA A100 GPU was used for training. All evaluations were performed under identical conditions, including the same encoder, data splits, and official codebases. Additional details and extended results are provided in the supplementary material.


\subsection{Cancer Subtyping}
\label{subtyping}
We consider two subtyping tasks \cite{peled2024multi}: MSS vs. MSI in Colorectal Cancer (CRC) and CIN vs. GS in Stomach Cancer (STAD), using datasets from TCGA\footnote{\href{https://portal.gdc.cancer.gov}{https://portal.gdc.cancer.gov}}.
We compared PSA-MIL to several contextual and non-contextual MIL baselines, concentrating on recent advancements in spatial modeling, such as SM-MIL \cite{castro2024sm}, Bayes-MIL \cite{cui2023bayes}.
\cref{main_table} presents the results.
Our PSA-MIL variants consistently outperform prior works across most metrics, demonstrating the effectiveness of learned spatial priors.
Notably, non-contextual methods perform competitively compared to previous contextual approaches, with IBMIL even surpassing certain contextual models.  
This suggests prior contextual methods may capture inaccurate spatial relationships, leading to suboptimal performance.
Among contextual methods, SM-MIL and Bayes-MIL performed well, highlighting the benefits of local smoothness and probabilistic modeling, respectively.
Furthermore, we observe performance differences among PSA-MIL variants, highlighting the impact of spatial distribution modeling on performance.

\subsubsection{Spatial Context Integration in Self-Attention MIL}
To further demonstrate the effectiveness of our approach, we compare it to several existing methods for integrating spatial context into self-attention MIL (MSA-MIL), described in \cref{self_attention_MIL}. 
Specifically, we benchmark against five baselines: the naive approach (with no spatial context), K-local attention (where only instances within the K-neighborhood are included in the self-attention, each with equal contribution), PPEG used in TransMIL \cite{shao2021transmil}, Relative Positional Encoding (RPE) suggested by SETMIL \cite{zhao2022setmil}, and RoPe 2D adaptation \cite{su2024roformer,pochet2023roformer}.

\noindent \cref{RPE_table} presents the results. PSA-MIL achieved the best results on both datasets, surpassing the baseline by +5.5\% in TCGA-CRC and +3.1\% in TCGA-STAD. 
RoPE showed competitive gains on TCGA-CRC but performed poorly on TCGA-STAD. Prior work has noted that RoPE may underperform when sequence lengths vary \cite{liu2023scaling,chen2023longlora}, and its 2D adaptation to pathology likely amplifies this limitation.
Other positional encoding methods provided some improvement, but their impact was significantly lower than ours, with RPE proving more effective than PPEG. 
The K-local attention variants yielded minimal to no performance gains, highlighting the limitations of fixed local constraints and the advantages of dynamic spatial modeling.

\begin{table}[h]
\centering
\resizebox{\linewidth}{!}{
\begin{tabular}{lccccccc}
\toprule
 & \shortstack{MSA-MIL\\$_{(baseline)}$} 
 & \shortstack{Local\\$_{[K=7]}$} 
 & \shortstack{Local\\$_{[K=15]}$} 
 & \shortstack{PPEG\\$ $} 
 & \shortstack{RPE\\$ $}  
 & \shortstack{RoPe\\$ $}  
 & \shortstack{PSA-MIL\\$_{[Gau]}$} \\

\midrule
TCGA-CRC  & 84.34 & 83.98 & 84.89 & 86.00 & 86.68 & 87.38\textsuperscript{*} & 88.97\textsuperscript{*} \\
\multicolumn{1}{c}{$\triangle$} 
          & -- 
          & {\color{red}\small$-0.4\%$} 
          & {\color{green!60!black}\small$+0.7\%$} 
          & {\color{green!60!black}\small$+2.0\%$} 
          & {\color{green!60!black}\small$+2.8\%$}
          & {\color{green!60!black}\small$+3.6\%$}
          & {\color{green!60!black}\small$+5.5\%$} \\
\midrule
TCGA-STAD & 80.52 & 80.80 & 81.19 & 81.25 & 81.91 & 81.06 & 82.98 \\
\multicolumn{1}{c}{$\triangle$} 
          & -- 
          & {\color{green!60!black}\small$+0.3\%$} 
          & {\color{green!60!black}\small$+0.8\%$} 
          & {\color{green!60!black}\small$+0.9\%$} 
          & {\color{green!60!black}\small$+1.7\%$}
          & {\color{green!60!black}\small$+0.7\%$}
          & {\color{green!60!black}\small$+3.1\%$} \\
\bottomrule
\end{tabular}}
\caption{AUC scores and improvements over the MSA-MIL baseline, comparing different spatial modeling strategies. Significant gains over baseline (paired t-test, $p < 0.05$) are marked with *.}

\label{RPE_table}
\end{table}

\subsection{Metastatic Detection}
\label{camelyon}
We evaluate our method on cancer metastatic detection using the CAMELYON16 dataset \cite{bejnordi2017diagnostic}, assessing both slide-level classification and patch-level localization performance. 
Patch-level scores are extracted and compared to ground-truth tissue annotations only at test time. See the supplementary materials for more details.
Results in \cref{combined_camelyon16_table} show that PSA-MIL achieved top slide-level classification performance, with AUC exceeding 96\% and the highest F1 score. Crucially, it also excelled in localization, achieving the highest AUC-FROC and AUC-ROC by a clear margin. 
PSA-MIL stands out by maintaining high performance across both tasks, highlighting its enhanced robustness compared to other models.

\begin{table}[h]
    \centering
    \resizebox{\linewidth}{!}{
    \begin{tabular}{lccccc}
        \toprule
        & \multicolumn{2}{c}{Slide-level} & \multicolumn{3}{c}{Patch-level} \\
        \cmidrule(lr){2-3} \cmidrule(lr){4-6}
        
        Method & \shortstack{AUC-\\ROC}  & \shortstack{F1 \\ $ $ } & \shortstack{AUC-\\FROC} & \shortstack{AUC-\\ROC} & \shortstack{F1 \\ $ $ }  \\
        \midrule
        ABMIL \cite{ilse2018attention}     & 95.3 & 91.5 & 67.8 & 88.8 & 66.6 \\
        DTFD-MIL \cite{zhang2022dtfd}   & \underline{96.1} & \underline{92.2} & 68.3 & 88.1 & 65.9 \\
        TransMIL \cite{shao2021transmil}   & 94.8 & 90.5 & 67.7 & 84.6 & 21.1 \\
        GTP \cite{zheng2022graph}       & 91.7 & 90.1 & 69.9 & 79.6 & 46.1 \\
        SM-MIL \cite{castro2024sm}    & \textbf{96.8} & 89.6 & 69.9 & 91.6 & 64.6 \\
        BAYES-MIL \cite{cui2023bayes}  & 95.0 & 91.8 & \underline{72.5} & \underline{93.6} & \textbf{76.5} \\
        \rowcolor{lightred}
        PSA-MIL\textsubscript{[Gau]} & \underline{\cellcolor{lightred} 96.1} & \textbf{\cellcolor{lightred} 92.3} & \textbf{75.9} & \textbf{94.7} & \underline{75.4} \\
        \bottomrule
    \end{tabular}}
    \caption{Slide-level classification and patch-level localization performance on CAMELYON16. 
    PSA-MIL achieved top slide-level performance while maintaining superior localization quality.}
    \label{combined_camelyon16_table}
\end{table}

\begin{figure}[h]
    \centering
    \includegraphics[width=\linewidth]{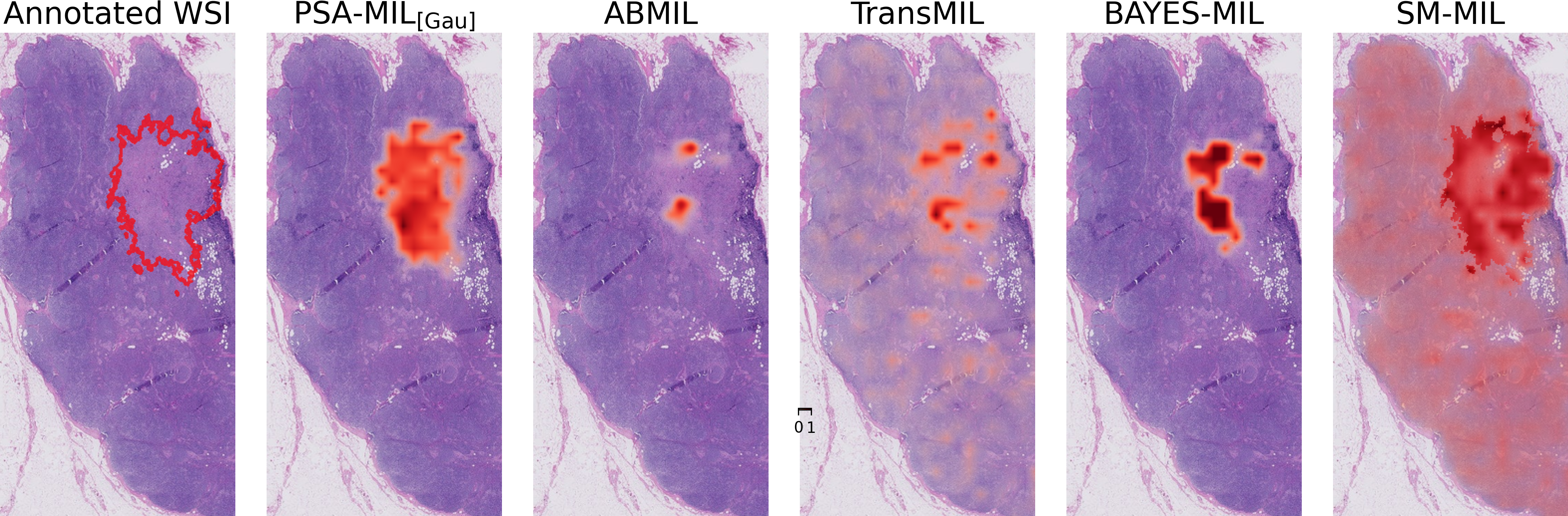}
\caption{Visual comparisons of different ROIs produced by various MIL models for metastatic detection. PSA-MIL accurately highlights the ROI; other methods show imprecise localization.
}
    \label{visual_compare_fig}
\end{figure}

\subsubsection{Visual Comparisons}
\label{visual_comparisons}
The quantitative improvements in localization are further illustrated in \cref{visual_compare_fig}, which shows heatmaps generated by different MIL baselines. 
PSA-MIL accurately spots the metastatic region, presenting a visually compelling and focused attention map. 
BAYES-MIL exhibited a more dichotomous attention distribution, with some regions of the metastatic area overlooked. SM-MIL, on the other hand, highlighted the entire ROI but produced a noisier attention map overall, likely due to over-smoothing. 
These findings emphasize PSA-MIL’s ability to generate precise and interpretable attention maps.

\subsection{Survival Prediction}
\label{survival}
We evaluated our approach for survival prediction using TCGA-CRC, TCGA-STAD, and TCGA-BRCA cohorts, assessing performance with the concordance index (c-Index).
We employed the survival cross-entropy loss as proposed in \cite{zadeh2020bias}.
Results using UNI encoder \cite{chen2024towards} are presented in \cref{surv_table}, with additional results using Lunit \cite{kang2023benchmarking} and GigaPath \cite{xu2024whole} encoders provided in the supplementary materials.

\noindent PSA-MIL\textsubscript{[Gau]} achieved significantly better results on TCGA-CRC and TCGA-STAD, as well as the second-highest performance on TCGA-BRCA. Other contextual methods, such as Sm-MIL and BayesMIL, failed to achieve competitive scores. Surprisingly, ABMIL demonstrated enhanced performance, outperforming several of the contextual methods. 
These results remained consistent across encoders, highlighting the robustness of PSA-MIL.

\begin{table}[ht]
    \centering
    \resizebox{\linewidth}{!}{
    \begin{tabular}{lcccccc}
        \toprule
        Dataset & AB- & DTFD- & Trans- & Sm- & Bayes- & PSA- \\
                &  MIL     & MIL   & MIL    & MIL & MIL    & MIL\textsubscript{[Gau]} \\
        \midrule
        TCGA-CRC 
            & 63.8 & 55.5 & 59.9 & 63.9 & 55.9 & \textbf{70.7} \\
        TCGA-STAD 
            & 58.6 & 53.3 & 52.9 & 56.6 & 52.4 & \textbf{61.1} \\
        TCGA-BRCA 
            & \textbf{61.9} & 53.9 & 57.8 & 57.2 & 49.9 & 60.2 \\
        \bottomrule
    \end{tabular}}
    \caption{
    C-index results for survival prediction using the UNI encoder \cite{chen2024towards}. Results with Lunit \cite{kang2023benchmarking} and GigaPath \cite{xu2024whole} (see supplementary) show consistent trends, confirming PSA-MIL's superiority across encoders.
    }
    \label{surv_table}
\end{table}

\subsection{Analysis}

\subsubsection{Computational Cost Analysis}
\label{computation}
\begin{figure}[t]
\centering
\includegraphics[width=0.9\columnwidth]{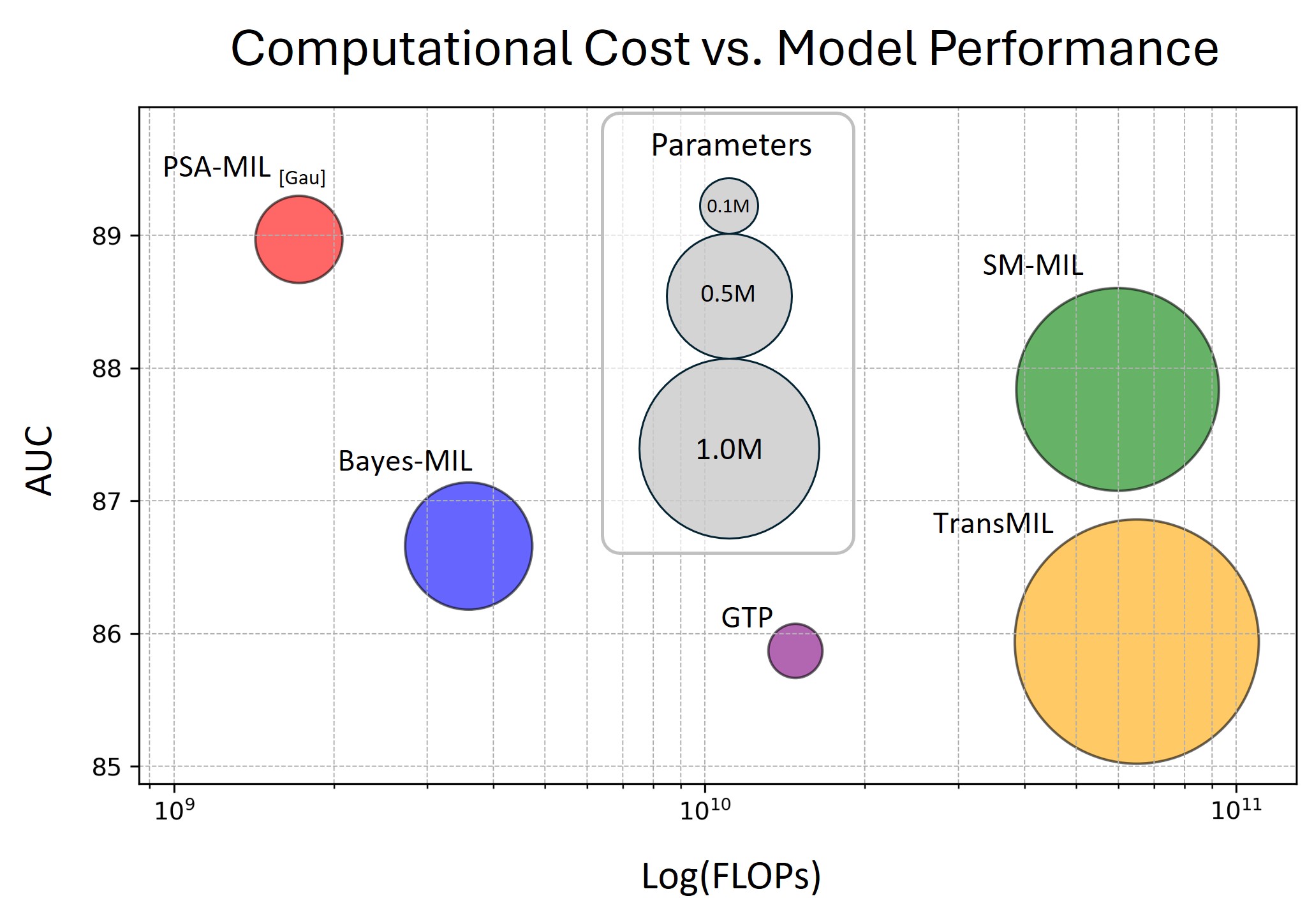}
\caption{Subtyping performance (AUC) vs. mean FLOPs per batch during training (log scale) for contextual MIL models on TCGA-CRC, with bubble size representing parameter count. PSA-MIL delivered the best performance with a notably smaller computational footprint.}
\label{comp_fig}
\end{figure}

Contextual MIL models inherently incur higher computational costs due to the need to model spatial interactions.
To further demonstrate the superiority of our approach, we conduct a computational cost analysis in \cref{comp_fig}, comparing AUC performance against FLOPs, with bubble sizes indicating the number of trained parameters. We report mean FLOPs per batch during training, providing a consistent measure of computational cost across models.

\noindent PSA-MIL achieves the highest AUC while maintaining the lowest FLOP count and fewer parameters, indicating the effectiveness of our proposed spatial pruning strategy. 
SM-MIL showed good performance (\cref{main_table}) but requires an order of magnitude more computational resources. 
GTP has a small parameter footprint yet suffers from high FLOPs and inferior performance.

\subsubsection{Analyzing Training Dynamics} 
\label{training_dynamics}
We analyze the training dynamics of PSA-MIL$_{[Gau]}$ on TCGA-CRC subtyping, focusing on the impact of diversity loss and the learned spatial attention patterns.
\cref{div_fig} (a-b) examines the evolution of our \textit{dynamic local attention} formulation. Without diversity loss, all heads converge to similar locality values, whereas with diversity loss, they exhibit more distinct behaviors, ensuring diverse spatial representations. Similarly, inter-head similarity remains relatively high without diversity loss (\cref{div_fig}c), while diversity loss (\cref{div_fig}d) encourages head specialization by reducing pairwise similarity.  
Additional visualizations of the attention heatmaps of PSA-MIL$_{[Gau]}$ are provided in the supplementary materials. These illustrate how different attention heads learn distinct spatial patterns. For example, the head with \(K{=}4\) produces highly localized activation spots, while the head with \(K{=}17\) captures broader, smoother attention distributions.

\begin{figure}
\centering
\includegraphics[width=1\linewidth]{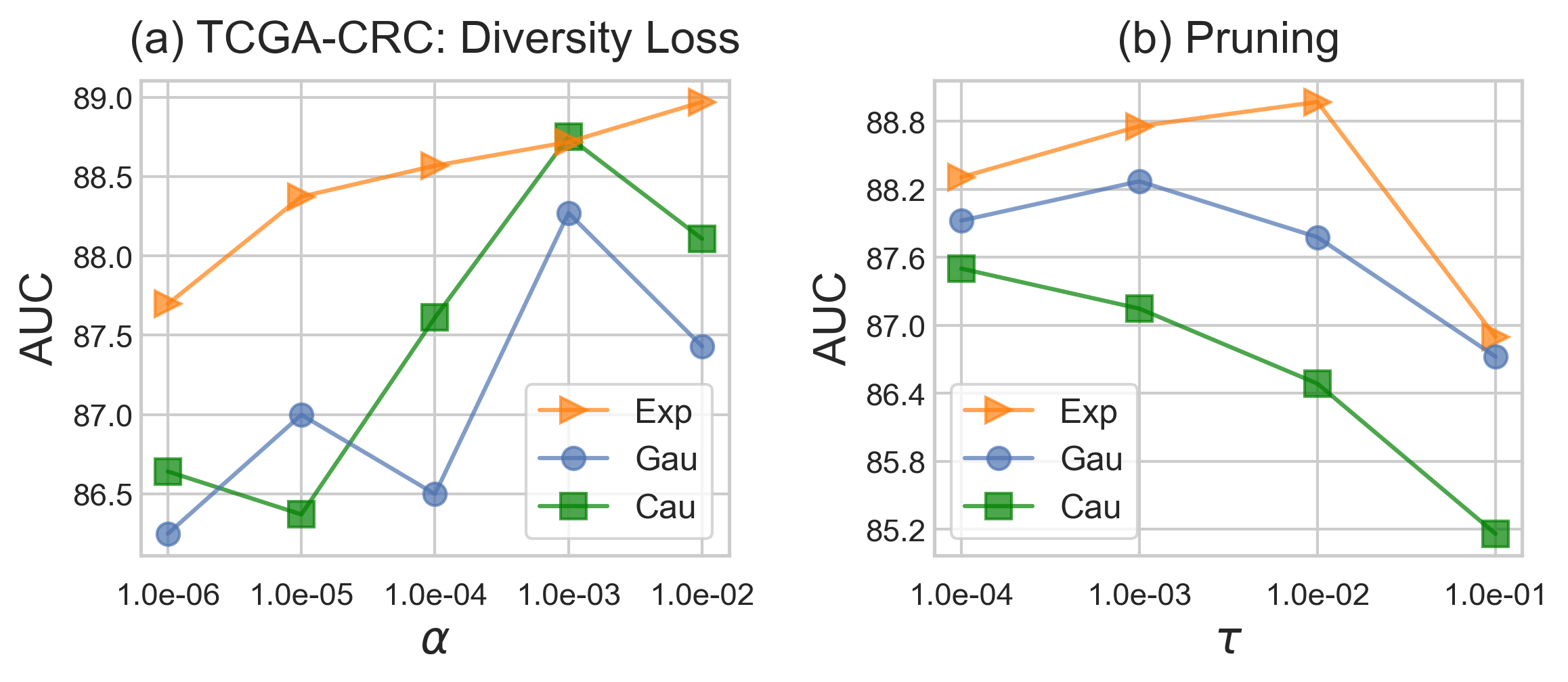}
\caption{Ablation studies on the impact of diversity loss (\(\alpha\)) and pruning threshold (\(\tau\)) in PSA-MIL on TCGA-CRC. Extended results are provided in the supplementary material.}

\label{ablation_fig}
\end{figure}


\subsection{Ablation Studies}
We report ablations on diversity loss weight (\(\alpha\)) and pruning threshold (\(\tau\)) in Fig.~\ref{ablation_fig}.
Ablations indicate that higher \(\alpha\) values generally improve performance, highlighting the beneficial effect of diverse representations.
Results remain relatively stable across \(\tau\), especially with Gaussian decay, enabling notable efficiency gains with minimal losses.
Full analysis with additional results is provided in the supplementary.




\section{Discussion}
PSA-MIL introduces a probabilistic interpretation of self-attention that incorporates spatial context through learnable distance-decayed priors, an entropy-based diversity loss, and a spatial pruning strategy. This formulation enables attention heads to independently capture diverse locality ranges, improving both interpretability and computational efficiency.
While effective, PSA-MIL is not without limitations. First, it remains bound to pretrained encoders, which can act as information bottlenecks if not chosen appropriately. 
Second, cross-domain evaluation, while highly valuable, is beyond the scope of this work and represents an important direction for future research.
In summary, PSA-MIL combines strong performance with efficiency, paving the way for broader applicability.

\section{Acknowledgments}
M.F.\ is supported in part by research grants from the Israeli Ministry of Science and Technology (1001577479, 1001577564), the Israel Innovation Authority (80783), and the Zimin Institute for AI Solutions in Healthcare.

\noindent Y.E.M.\ is supported in part by the Israeli Science Foundation personal grant 2794/21 and the Israeli Personalized Medicine Program grant 3120/22.

{
    \small
    \bibliographystyle{ieeenat_fullname}
    \bibliography{main}
}

\end{document}